\documentclass[preprint]{elsarticle}

\journal{SI of Robotics and Autonomous Systems}

\usepackage{amssymb}
\usepackage{bm}
\usepackage{amsmath}
\usepackage{rotating}
\usepackage{color}
\usepackage{algorithm} 
\usepackage{algorithmic}
\usepackage{soul}
\usepackage{changes}
\usepackage[hidelinks]{hyperref}
\usepackage{doi}

\begin{document}
\definechangesauthor[color=magenta]{AT}

\begin{frontmatter}

\title{Learning agent's spatial configuration\\from sensorimotor invariants}
\tnotetext[mytitlenote]{This work was conducted within the French/Japan BINAAHR project under Contract n◦ ANR-09-BLAN-0370-02 funded by the French National Research Agency. J.K.O'Regan \& A.V.Terekhov were financed by ERC Advanced Grant 323674 ``FEEL''.}

\author[UPMC,ISIR]{Alban Laflaqui\`ere}
\ead{alban.laflaquiere@gmail.com}

\author[LPP]{J. Kevin O'Regan}
\ead{jkevin.oregan@gmail.com}

\author[UPMC,ISIR]{Sylvain Argentieri}
\ead{sylvain.argentieri@upmc.fr}

\author[UPMC,ISIR]{Bruno Gas}
\ead{bruno.gas@upmc.fr}

\author[LPP]{Alexander V. Terekhov\corref{mycorrespondingauthor}}
\cortext[mycorrespondingauthor]{Corresponding author}
\ead{avterekhov@gmail.com}

\address[UPMC]{Sorbonne Universit\'es, UPMC Univ Paris 06, UMR 7222, ISIR, F-75005, Paris, France}
\address[ISIR]{CNRS, UMR 7222, ISIR, F-75005, Paris, France}
\address[LPP]{Laboratoire Psychologie de la Perception, Universit\'e Paris Descartes, CNRS and \'Ecole Normale Sup\'erieure}

\begin{abstract} The design of robotic systems is largely dictated by our purely human intuition about how we perceive the world. This intuition has been proven incorrect with regard to a number of critical issues, such as visual change blindness. In order to develop truly autonomous robots, we must step away from this intuition and let robotic agents develop their own way of perceiving. The robot should start from scratch and gradually develop perceptual notions, under no prior assumptions, exclusively by looking into its sensorimotor experience and identifying repetitive patterns and invariants. One of the most fundamental perceptual notions, space, cannot be an exception to this requirement. In this paper we look into the prerequisites for the emergence of simplified spatial notions on the basis of a robot's sensorimotor flow. We show that the notion of space as environment-independent cannot be deduced solely from exteroceptive information, which is highly variable and is mainly determined by the contents of the environment. The environment-independent definition of space can be approached by looking into the functions that link the motor commands to changes in exteroceptive inputs. In a sufficiently rich environment, the kernels of these functions correspond uniquely to the spatial configuration of the agent's exteroceptors. We simulate a redundant robotic arm with a retina installed at its end-point and show how this agent can learn the configuration space of its retina. The resulting manifold has the topology of the Cartesian product of a plane and a circle, and corresponds to the planar position and orientation of the retina.  \end{abstract}


\end{frontmatter}

\section{Introduction}
\label{sec:Introduction}

The classical approach to the control of robotic systems consists in developing an electro-mechanical model of the robot, defining the range of operating conditions, and building algorithms for robot/environment state estimation and robot control. This approach, while efficient in highly controlled environments (such as robotized factories), can lead to severe failures when operating under unforeseen circumstances. Making robots more autonomous and capable of operating in unstructured a priori unknown environments is the main objective of modern robotic research.

The application of techniques from machine learning and artificial intelligence has enabled significant progress in this direction over recent decades. Numerous solutions have been proposed for online robot self-modelling~\cite{Sigaud2011, Hersch2008, Sturm2009, Yoshikawa2004, Martinez2010} (see also \cite{Hoffmann2010} for a comprehensive review) and recovery from unknown damages~\cite{Bongard2006, Koos2013, Wiesel2005}. These solutions usually define a set of building blocks (such as rigid bodies) and the rules of their connection (e.g. joints) and try to find the combination of these blocks that best accounts for the incoming sensory information given the motor commands. Examples using such an approach can be found for instance in \cite{Tsai1988,Bennett1991} and more recently in \cite{Schilling2011, Hersch2009}. Another approach avoids defining building blocks but instead adds some pre-processing of the sensory flow to avoid facing its raw complexity, generating inputs that suit the task. Such pre-processing can for instance be used to define the coordinates of the robot's hand in the visual field to learn a kinematic model of the arm \cite{Chao2014, Jamone2012} or a target object for reaching \cite{NataleMetta2007, Gaskett2003}.

Although this type of approach often produces spectacular results, its robustness and efficiency strongly depend on the choice of the building blocks and pre-processing algorithms. This makes the entire approach heavily biased by the designer's intuition, which is rooted in human perception and which is not necessarily best suited for robots, whose sensors and effectors differ significantly from those of human beings. To understand the implications of this difference, consider the fact that the entire field of computer vision has been biased by the false perceptual intuition that seeing is similar to having a photo of the visual scene~\cite{Rensink1997,ORegan1999} and that stepping aside from this paradigm can yield unexpectedly fruitful results~\cite{Benosman2011,Lorach2012,Rogister2012}. Thus, to make robotic systems truly autonomous and robust, we need to shed the biases imposed on us by our own perceptual system and let robots develop their own ways of perceiving the world. Few studies adopting such a radical approach to control robots have been proposed \cite{Censi2012, Hoffmann2013, Kuipers2008}. Although these studies are in line with our approach, they do not directly address the problem of space perception (or only implicitly through the robot's ability to move in its environment). This will be the main focus of the present paper.

In order to minimize a prioris about perceptual systems we consider a robotic agent designed as a \emph{tabula rasa} receiving undifferentiated sensory inputs (e.g. not knowing whether a given one of them comes from a video camera or from an encoder in a joint) and sending out undifferentiated motor outputs. Perceptual structures can emerge as stable patterns in the agent's sensorimotor flow. This approach, when applied to visual information, can lead to the discovery of stable features, such as edges, similar to those present in the human visual cortex~\cite{Schmidhuber1996,Olshausen1996,Masquelier2007,Lee2008,Choe2008}. A similar approach, formalized in the language of information theory~\cite{Klyubin2004}, can make it possible to describe the topological structure of the agent's surface~\cite{McGregor2011} and certain properties of its interaction with the environment~\cite{Kaplan2004,Klyubin2005a,Gordon2011}. The cited studies clearly show that \emph{tabula rasa} agents can learn basic properties of the available sensorimotor information. However, only the simplest perceptual notions are straightforwardly dictated by the sensory inputs themselves. More complicated notions represent laws linking sensations and actions, rather than particular instances of sensory information \cite{OReganNoe2001}. Space, which does not correspond to any particular sensory inputs, is one such a notion.

Can a \emph{tabula rasa} approach be used to model the acquisition of the notion of space? In order to answer this question we first need to decide on what we mean by the term \emph{space}. In mathematics, various spaces are described: topological, metric, linear, etc. Each of these notions captures certain features of what we usually mean by space. For example, topological spaces only feature the notion of proximity, and can be thought of as reflecting an agent with a highly impaired ability to make distance judgments (which is true for humans performing certain tasks). Although rather primitive, topological space nevertheless includes some fundamental aspects of space in general, such as dimensionality, and its notion can be useful in such tasks as the mapping of large spaces ~\cite{Pierce1997}. Metric spaces are more complicated objects, which imply precise information of distances. They provide the tool required to work with such notion as the length of a path, and can underlie navigation abilities. In particular, knowledge of a metric space enables odometry and SLAM~\cite{Mueller1988, Smith1990, Bowling2007}. Linear spaces introduce the notion of the vector, which is an efficient tool for describing motion. The link between motion and linear spaces is used in many studies that address the problem of space acquisition. Thus, Poincar\'e~\cite{Poincare1895} suggested that spatial knowledge emerges from the agent's capacity to move, with spatial relations such as the distance to an object being internally encoded as potential motor commands. The agent's ability to move has also played an essential role in more recent works on space \cite{Pierce1997, Stober2011, Roschin2011}. Philipona and co-authors showed in~\cite{Philipona2003} that under certain conditions the dimensionality of space can be estimated by analyzing only sensorimotor information that is available to the agent. This result launched a series of publications by the present authors, extending the conditions of dimension estimation \cite{Laflaquiere2012} and applying similar ideas to different agents and robotic systems \cite{Laflaquiere2010,Bernard2012,Laflaquiere2013}.

Knowing the number of spatial dimensions is not, however, the same as having the notion of space. It has recently been shown that the notion of space can be learned as a proprio-tactile mapping \cite{Roschin2011} or as a group of rigid transformations of the environment \cite{Terekhov2013}. Here we focus on a different aspect of spatial knowledge, probably the simplest that can be extracted by a naive agent. In order to introduce it, let us first note that mathematical spaces (topological, metric, etc.) do not emphasize what is special about our subjective experience of space. Mathematical spaces can be applied, for example, to describe the full set of an agent's body postures, or motor commands, or even the outputs of every pixel in the agent's visual sensor (e.g. camera). However, these examples clearly do not correspond to what we usually mean by space. We believe that what characterizes space is the particular structure that it imposes on possible sensorimotor experiences. It can be identified in the laws that govern the way sensory inputs change as the agent moves around.

The first and most basic property of those laws is an invariance: space does not depend on the particular environment, nor on the particular posture of the agent. The agent must somehow know that its sensor is at the same spatial position independently of what objects are around and what are the positions of the other sensors. In other words, the first aspect of the notion of space is the ``point of view'' from which the agent ``looks'' at the world (here we adopt visual terminology for simplicity, but the notion must not depend on the particular type of sensors in question: camera, microphone, or taxel array). From now on, when we speak about the notion of space we will be referring to the set of the agent's ``points of view''. These ``points of view'' are the precursors to the more convenient notion of ''point'', which is the basic element of what we call space, and which can be used to build more complex notions of space. Note that in our approach, we are looking at the problem of space from the agent's point of view. Instead of taking the existence of external space for granted, we are trying to identify signs of it in the agent's sensorimotor flow, and to see what makes the notion of space useful to the agent. Our hope is that by learning how the notion of space can be constructed from the sensorimotor flow we will acquire a better understanding of how other perceptual notions can be learned, such as body, object, etc. In this respect, our work is notably different from the field of research on body schema acquisition, which is already the subject of a large literature. The question of space is usually eluded in these studies, as it is supposed to be either an unnecessary prerequisite for action or to already have been acquired.

This study extends our previous work~\cite{Laflaquiere2013}, where a neural network was used to learn a mapping between the motor space and an internal representation of the agent's external configuration. This internal representation was generated online during the exploration of multiple environments. The present paper introduces two main improvements. First, it offers a clear definition of the structure of the constraints captured by the agent. In doing so, it makes explicit the mapping that was implicitly captured by the neural network in our previous study. Second, the metrics of the internal representation are no longer derived statistically from the exploration of multiple sensory manifolds, but from motor data. This ensures its independence from the actual content of the explored environments. The current study complements our previous work~\cite{Terekhov2013}, which assumes that the agent already possesses the notion of space as a collection of ``points of view'' and determines a group of rigid displacements over them, thus approaching the notion of geometry in the sense of Klein.

The paper is organized as follows. Section~\ref{sec:Part1} is dedicated to the problem of space acquisition, illustrated with a simplified agent. The representation variance w.r.t.\ the environmental changes is highlighted, and a solution to this problem is proposed. Next, it is shown in Section~\ref{sec:Part3} how the manifold of points of view can be obtained in a more complicated case, and a discussion on the genericity of the results follows. Finally, a conclusion ends the paper.

\section{Problem statement}
\label{sec:Part1}

In this section, two simple scenarios are presented to illustrate the key idea of the paper. A first agent is introduced to show why considering only exteroception is problematic when attempting to produce a stable internal representation of spatial properties. A solution to this problem is proposed through a second agent, with the introduction of motor information into the constraint-capturing process.

\subsection{Formalization and definitions}
\label{sec:Part1_0}
In the following, we consider so-called \emph{naive} agents which have no a priori knowledge about the world they are immersed in and, in particular, about the existence and structure of space. We assume that they have access to both motor and exteroceptive sensory information. We also assume here that any possible proprioceptive sensory information is a redundant copy of motor information. We  thus ignore it and consider only motor information.
The agent's motor output and sensory input are respectively defined as:
\begin{equation}
	\mathbf{m} = (m_1,m_2,\dots,m_{N})^T \text{ ; } \mathbf{s} = (s_1,s_2,\dots,s_K)^T,
\end{equation}
where $m_n, n \in \{1,\dots,N\}$ is the command driving the $n$-th motor and $s_k, k \in \{1,\dots,K\}$ is the output of the $k$-th sensor distributed on the agent's body. The sets of all possible outputs $\mathbf{m}$ and inputs $\mathbf{s}$ are respectively called \emph{motor} and \emph{sensory space}. 

For any state of the environment, denoted $\bm{\mathcal{E}}$, we postulate a sensorimotor law $\sigma$ such that:
\begin{equation}
\mathbf{s} = \sigma_{\bm{\mathcal{E}}}(\mathbf{m}).
\label{eq:sigma}
\end{equation}
This relation specifies that, according to the physics of the world, any motor output $\mathbf{m}$ is associated with a sensory input $\mathbf{s}$ for a given environment $\bm{\mathcal{E}}$. Note that the sensorimotor law is of course unknown to the naive agent. 

The objective of this paper is to understand why and how the notion of space can emerge in such a naive agent whose whole experience is captured by this unknown sensorimotor law. From an external point of view, space is usually understood as a container in which the agent and objects are immersed and in which they are endowed with the notions of (relative) position/orientation and can go through transformations (displacements). Nonetheless, explaining how such sophisticated notions can emerge from an agent's raw sensorimotor interactions with its environment(s) is far from trivial. In particular, we need to be able to answer the following question: \emph{how can the agent discover that there exists a stable external structure within which it lives and which corresponds to what we, humans, call ``space''?}

From the point of view of a naive agent with no a priori knowledge about space, this notion needs to be reconsidered through the prism of sensorimotor information. We propose to consider space as \emph{a set of constraints that all possible functions $\sigma_.(.)$ satisfy}. Moreover, since the structure of space should not depend on its content, \emph{those constraints have to be invariant to the environmental state $\bm{\mathcal{E}}$}.
In this paper, the naive agent's objective is to capture the existence of these constraints. More precisely, we assume that the agent's intrinsic drive is to capture structures underlying its sensorimotor experience and to represent it in an economical way. We hypothesize that the discovery of the spatial constraints can be materialized by the construction of an internal representation of the agent's configuration with respect to the external space it is immersed in. It can be argued that such an explicit internal representation is unnecessary. As suggested in~\cite{Laflaquiere2013}, the constraints could of course be captured implicitly in the system. The explicit internal representation will however be useful for an algorithmic / semantic agent, as opposed to a distributed neural-networks-based agent. In what follows, this approach is illustrated with two ``toy examples'' and then applied to a simulated robot arm in section~\ref{sec:Part3}.

\subsection{The variability of sensory experience}
\label{sec:Part1_1}
When thinking about space, there is a natural tendency to focus on information provided by the agent's exteroceptive flow. Without any a priori knowledge, this is indeed the only source of information about the external world. In keeping with this fact, most work in robotics about the discovery of spatial properties relies on the analysis of exteroceptive data. One example~\cite{Wyss2006} can be cited where a camera is used to infer the approximate topology of the environment that a mobile robot operates in. In an overwhelming majority of cases, exteroceptive sensory data are even pre-processed so as to already carry spatial knowledge according to the roboticist's own a priori knowledge (for instance the position of the robot's hand in~\cite{Sturm2009, Cheah2010}). However, as implied by \eqref{eq:sigma}, the $K$ raw sensory inputs $s_k$ depend on the state of the environment $\bm{\mathcal{E}}$: the sensory input $\mathbf{s}$ changes with the objects in the environment. Thus, it is not obvious how the notion of space, which should be independent of a particular environment, can be extracted from data that depends so heavily on it.
\begin{figure}[tb]
\centering
\includegraphics[width=.7\linewidth]{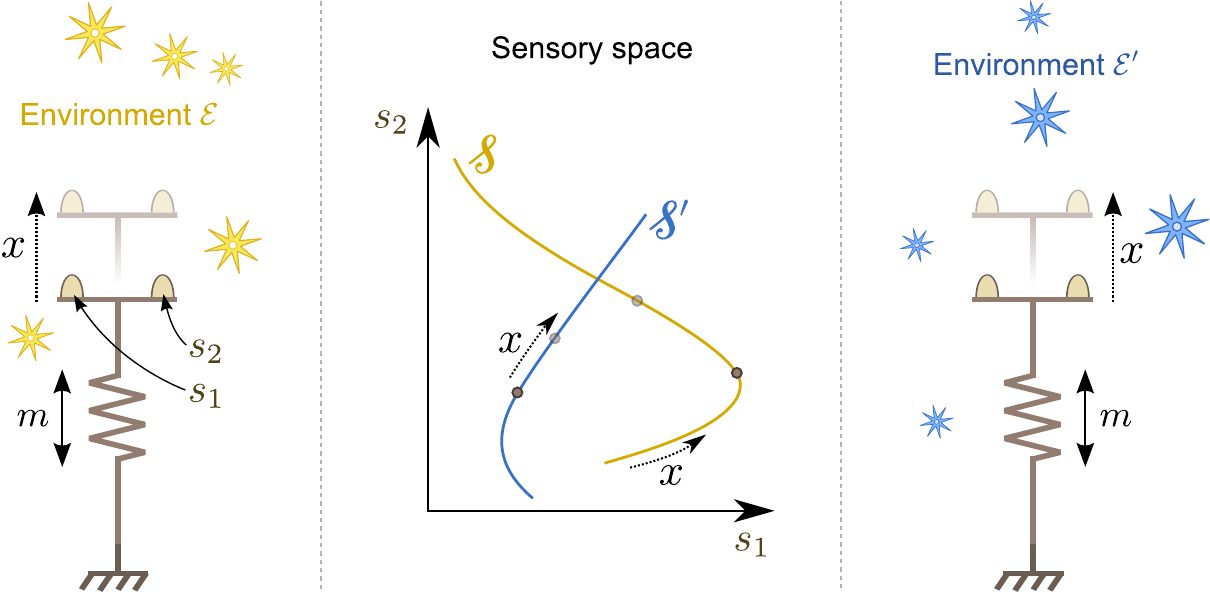}
\caption{The agent can move its sensors in external space using its motor. Although the external agent configuration $x$ can be the same, its sensory experience varies greatly depending on the structure of the environment.}
\label{fig:ToyExample1}
\end{figure}

To illustrate the problem, consider the so-called ``toy agent'' depicted in Fig.~\ref{fig:ToyExample1}. For the sake of simplicity, the agent is immersed in a $2$D space. It has a simple T-shaped structure which is moved along a line by a single motor command $\mathbf{m}=m_1$. It is endowed with two light sensors placed on its top, such that $\mathbf{s}=(s_1,s_2)^T$. The environment is made up of $L=5$ point light sources randomly distributed around the agent. The sensory inputs are generated as follows:
\begin{equation}
	s_i = \sum_{l=1}^{L} 1/d_{i,l}^2, i=\{1,2\},
\end{equation}
where $d_{i,l}$ is the Euclidean distance between the $i$-th sensor and the $l$-th light source.

The agent can explore its environment by moving along a line (parametrized by $x$ from an external point of view), for example by sending out random values of the motor command $\mathbf{m}$. For each motor output, it receives the corresponding sensory input $\mathbf{s}$.
If the environment is static and invariant during exploration, these collected outputs form a $1$D manifold $\mathcal{S}$ in the agent's sensory space (see Fig.~\ref{fig:ToyExample1}, center). This result was expected, as the dimensionality of the manifold $\mathcal{S}$ is determined by the number of degrees of freedom during exploration \cite{Laflaquiere2010}.
In some sense, the sensory manifold $\mathcal{S}$ describes the agent's internal experience of its translation in space, and is the only information it can gather about space.
However, $\mathcal{S}$ depends dramatically on the environment: the same exploratory movements performed in a different environment $\bm{\mathcal{E}'}$ will yield another sensory manifold $\mathcal{S'}$, which is totally different from $\mathcal{S}$ (see Fig.~\ref{fig:ToyExample1}, center and right). Consequently, it is difficult for the agent to extract knowledge about space --- which should be independent of the state of the environment --- from exteroceptive experience.

\subsection{Introducing motor information to capture external information}
\label{sec:Part1_2}

Biological findings suggest that the development of visual spatial perception is extremely dependent on the agent's ability to move actively, and not simply observe the visual changes caused by passive displacements in space~\cite{Held1963}. A similar conclusion can be drawn from sensory substitution experiments~\cite{Bach-y-Rita:2003it}. This difference between the active and passive generation of the same sensory experience reflects the importance of the knowledge of motor commands during the acquisition of spatial knowledge. From a more philosophical point of view, Poincar\'e~\cite{Poincare1895} also suggested that the notion of space and displacements should necessarily be rooted in the agent's ability to move. We thus propose to introduce motor information in our attempt to establish how a naive agent can discover space.

As described above, we define space as a set of constraints on the sensorimotor relationship $\mathbf{s}=\sigma_{\bm{\mathcal{E}}}(\mathbf{m})$ which maps motor output $\mathbf{m}$ onto the sensory space. It thus seems natural to look for space not in exteroceptive data themselves, but in the laws linking motor actions $\mathbf{m}$ to the sensory inputs $\mathbf{s}$. Moreover, since space should be independent of its content, some of the constraints on these laws have to be invariant to the state of the environment.

\begin{figure}[tb]
\centering
\includegraphics[width=.7\linewidth]{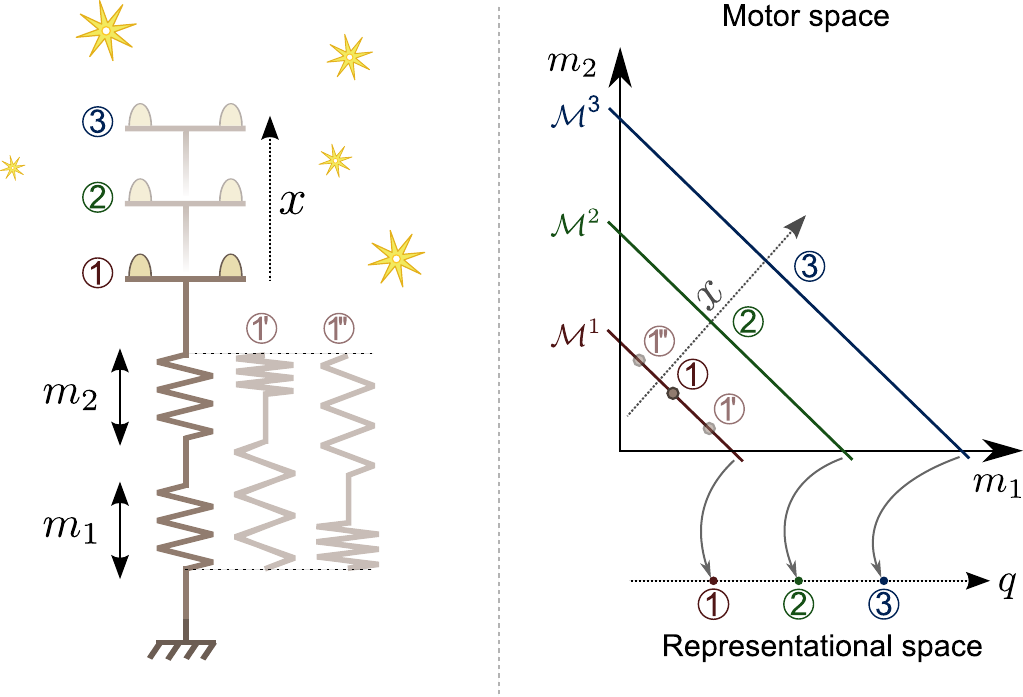}
\caption{The agent can move its sensors using two redundant motors. Any external configuration $x$ of the sensors is thus internally related to a set of redundant motor configurations $\mathcal{M}^i$. Those sets define a manifold which can be used to represent the external configuration via an internal parameter $q$.}
\label{fig:ToyExample2}
\end{figure}

To illustrate the process by which this invariance can be captured, let us introduce a slightly less trivial toy agent which differs from the previous one in that it has two redundant actuators, $m_1$ and $m_2$, as shown in Fig.\ref{fig:ToyExample2}. Just like the previous one, this new agent can explore its environment by sending random commands $\mathbf{m}=(m_1,m_2)^T$ and moving its exteroceptors in space along what, from an external point of view, is the same line. Evidently, for the environment $\bm{\mathcal{E}}$ the second agent will discover the same sensory manifold $\mathcal{S}$ as the first (see Fig.\ref{fig:ToyExample1}, center). However, the structure of its motor space is different, since $2$ motors are now involved in the generation of the $1$D manifold $\mathcal{S}$.
In other words, there exist distinct sets of motor outputs $\mathbf{m}$ that generate the same sensory input $\mathbf{s}$.
Hence, for any sensory input $\mathbf{s}$ from $\mathcal{S}$ related to a position of the agent in space, there exist a variety of motor outputs $\mathbf{m}$ such that $\sigma_{\bm{\mathcal{E}}}(\mathbf{m}) = \mathbf{s}$, for a given environmental state $\bm{\mathcal{E}}$. 
Such a coincidence in sensory inputs can be noticed, and the underlying motor outputs can be associated, defining the sets
\begin{equation}
\mathcal{M}^{\mathbf{s}} = \left\{\mathbf{m} \mid \mathbf{s} = \sigma_{\bm{\mathcal{E}}}(\mathbf{m})\right\}, \forall \mathbf{s} \in \mathcal{S}.
\end{equation}
The sets $\mathcal{M}^{\mathbf{s}}$ can be seen as the kernels of the functions $\sigma_{\bm{\mathcal{E}}}-\mathbf{s}, \forall \mathbf{s} \in \mathcal{S}$.

The aforementioned structural difference between the motor space and the sensory manifold $\mathcal{S}$ reveals the existence of constraints that apply to sensorimotor experience. Assuming an intrinsic drive to reduce the complexity of the internal representation of the agent's sensorimotor experience, the agent should thus be driven to capture the structure underlying those constraints. By doing so, it discovers the notion of space, as we defined it above. Such a discovery would not be expected in the first toy agent. Indeed, no additional structure was required to capture the sensorimotor experience, as there was a one-to-one relationship between motor output $\mathbf{m}$ and the corresponding sensory state $\mathbf{s}$. Rigorously speaking, it can be argued that a notion of space was also captured in the first toy example, as in that case too space exists and constrains the sensorimotor system. But those constraints are so trivial that they do not stimulate the agent to capture any underlying structure in its experience. It thus does not fit our definition.

For the example under consideration, the two motor commands $m_1$ and $m_2$ are redundant in determining the agent's position $x$ in space . The sets $\mathcal{M}^\mathbf{s}$ thus take the form of oblique lines, such that $m_1+m_2=x$ (see Fig.~\ref{fig:ToyExample2}, right).
In general and for any agent, if the environment $\bm{\mathcal{E}}$ is rich enough, the sensory input $\mathbf{s}$ is different for different positions of the exteroceptors in space. 
Note that in very peculiar environments this statement can be false. Seen through the agent's sensors, some environments could for instance present symmetries such that multiple spatial configurations of the agent would generate the same sensory input. If the same symmetry were to recur in every explored environment, the final structure captured by the agent would differ from what an external observer would expect. This is known to be the case for animals raised in particularly structured visual environments~\cite{Held1963, Blakemore1970}. However, it must be understood that to the agent's subjective experience, such an altered notion of space would be perfectly suitable as long as it worked for all environments that the agent encountered. On the other hand, if such symmetries are rare, we assume that the agent tries to capture a structure that is generic to all the explored environments. Such exotic cases are thus statistically filtered out in the experience accumulated during the agent's lifetime. Those particular cases aside, every set $\mathcal{M}^\mathbf{s}$ corresponds to a certain input $\mathbf{s}$ but also to a certain position of the agent in space. For a different environment $\bm{\mathcal{E}'}$, the sensory manifold can change to $\mathcal{S}'$ but the sets $\mathcal{M}^{\mathbf{s'}}$ will be exactly the same as the structure of the agent is unchanged and it is exploring the same external positions $x$. Even if sensations $\mathbf{s}$ depend on the environment, the sets $\mathcal{M}^\mathbf{s}$ are thus environment-independent. As a consequence, we will use the index $i$ instead of $\mathbf{s}$ to denote a particular set $\mathcal{M}^i$.

All the possible sets $\mathcal{M}^i$ have a specific structure that can be captured to build an internal representation of the external agent's configuration. Thus, each set $\mathcal{M}^i$ can be considered as a point on some new manifold. For this purpose, a metric function has to be introduced to define a distance $\rho(\mathcal{M}^j,\mathcal{M}^k)$ between every pair of sets $\langle \mathcal{M}^j,\mathcal{M}^k \rangle$. Together these distances define a manifold where each point is internally related to a motor set $\mathcal{M}^i$ and externally to an agent position $x$. It can therefore be used as an internal representation of the agent's external configuration.
For the second toy agent, the definition of a distance between each line $\mathcal{M}^i$ in the motor space (see Fig.\ref{fig:ToyExample2}, right) leads to the construction of a $1$D manifold. The parameter $q$ used to move along this manifold is an internal representation of the external parameter $x$.
Note that to obtain this internal representation the agent does not have to assume the existence of space: it simply has to associate the motor commands resulting in the same sensory input with each other. The notion of space is thus a byproduct of the agent's drive to capture its sensorimotor experience in a compact way.

This section has been devoted to the presentation of the key idea that \emph{despite the environment-dependency of exteroception, an invariant representation of the external agent's configuration can be built by taking into account motor information.} After a short illustration of this paradigm with two very basic scenarios, in the following section we use a more realistic agent to assess the approach.

\section{Application to a robotic arm}
\label{sec:Part3}

In this section, we simulate a robotic arm with several degrees of freedom to test the idea that spatial information can be acquired by looking into the redundant sets $\mathcal{M}^i$, which are the kernels of the function $\sigma_{\bm{\mathcal{E}}}-\mathbf{s}$. With minimal a priori knowledge, our objective is to build an invariant internal representation of the arm's end-point configuration to illustrate the capture of the constraints imposed by the existence of space. First we present the agent and the way sensory inputs are simulated. Second, the agent's exploration and the internal representation of its external configuration are described. Finally, the results are analyzed and discussed.

\subsection{Description of the agent}
\label{sec:Part3-1}
\begin{figure}[t]
\centering
\includegraphics[width=1\linewidth]{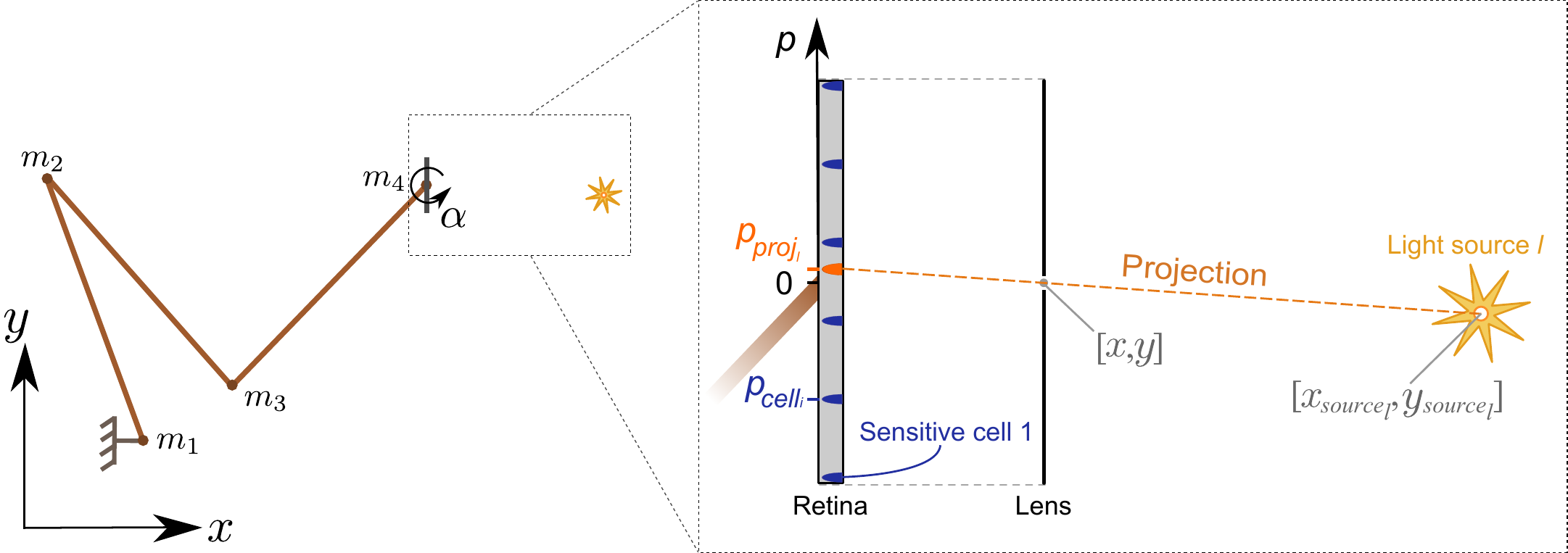}
\caption{On the left: the agent is a three-segment arm equipped with $4$ motors and a light-sensitive retina. On the right: the retina is regularly covered with $6$ light-sensitive cells. Light from the light sources is projected onto the retina through a pinhole lens. Each cell's excitation is a Gaussian function of its distance to the light projection.}
\label{fig:Pinehole}
\end{figure}

The agent is a three-segment serial arm with a retina-like sensor on its end-point (see Fig.\ref{fig:Pinehole}). Each segment is one unit long and each of the agent's $N=4$ hinge joints is actuated by a motor (with command $m_i$, $i\in \{1,\dots,4\}$). The agent thus has four motor outputs controlling the three external parameters of its retina: its $[x,y]$ position and orientation $\alpha$, although it has no knowledge of it.

The retina-like sensor is similar to a pinhole camera (see Fig~\ref{fig:Pinehole}). The retina is regularly covered with $6$ cells which are sensitive to light sources in the environment. The unitary excitation produced on the $i$-th cell by a punctual light source $l$ is:
\begin{equation}
 s_{i,l} = \frac{\exp(-||p_{cell_i}-p_{proj_l}||^2)}{||[x,y]-[x_{source_l},y_{source_l}]||}
\end{equation}
with $p_{cell_i}$ being the retinal position of the $i$-th cell, $p_{proj_l}$ the retinal position of the $l$-th light source projection, $[x,y]$ the lens's position in external space and $[x_{source_l},y_{source_l}]$ the position of the $l$-th light source in external space.
The environment is made up of $L=10$ randomly distributed point light sources. The total excitation $s_i$ of the $i$-th cell is:
\begin{equation}
s_i = \sum_{l=1}^{L} s_{i,l}.
\end{equation}

\subsection{Algorithm for the generation of the internal representation}
\label{sec:Part3-2}
The internal representation of the retina's configuration in space is generated in three steps. The algorithm used during the simulation is presented in \ref{sec:Algorithm}. Note once again that building an explicit internal representation of the captured structure is not mandatory. This knowledge could be captured implicitly in the system, as in the neural network presented in~\cite{Laflaquiere2013}. However, having some explicit internal representation of the sets $\mathcal{M}^i$ and a metric defined on them is useful for an algorithmic / semantic system as opposed to a neural network system. For example, such a representation can be used to reduce the dimensionality when sampling a new visual environment: the agent with an explicit internal representation ``knows'' that it needs to sample the three-dimensional space of the internal representation and not the full four-dimensional space of motor commands. In addition to this, in what follows here we  build an illustration of this structure as a means to visualize and interpret the knowledge extracted by the agent.

\paragraph{Step 1}
The agent collects the sensorimotor data by sending random outputs $\mathbf{m}^i=(m_1^i,m_2^i,m_3^i,m_4^i)^T$ to its motors. To simplify the presentation of the results (see section \ref{sec:Discussion}), we only analyze configurations whose corresponding retinal position is located within a rectangular working space. The size of the working space was arbitrarily set to $\{2,1,5\}$ units (height and width) and with its center located $1.75$ units in front of the agent. The analysis is restricted to $2500$ different arm configurations. Figure Fig~\ref{fig:Arm_manifold}a presents the working space and a sampling of the $2500$ arm configurations.

\paragraph{Step 2} A sensory input $\mathbf{s}^i$ is associated with every motor output $\mathbf{m}^i$. The sets $\mathcal{M}^i$ of redundant motor outputs generating the same inputs $\mathbf{s}^i$ are then determined. They are estimated using the local Jacobian of $\sigma_{\bm{\mathcal{E}}}$ and following the direction of the kernel in its motor space (see details in Appendix~A). Note that the naive agent does not know this Jacobian, and that it is introduced here only to accelerate the simulation. A more realistic scenario, in which these manifolds are estimated purely on the basis of unlabelled sensorimotor data, is described in \cite{Laflaquiere2013}. A sampling of the $2500$ manifolds $\mathcal{M}^i$ corresponding to the outputs $\mathbf{m}^i$ is illustrated in Fig.~\ref{fig:Arm_manifold}b. Each manifold in the figure is approximated by 100 points.

\paragraph{Step 3} An internal representation of the external retina configuration is determined as the manifold of $\mathcal{M}$'s. This manifold is defined by computing the Hausdorff distance for every pair $\langle\mathcal{M}^i, \mathcal{M}^j\rangle$. Note that any other metrics could be used instead to define the distances between the sets $\mathcal{M}^i$.
\begin{figure}[t!]
\centering
\includegraphics[width=0.6\linewidth]{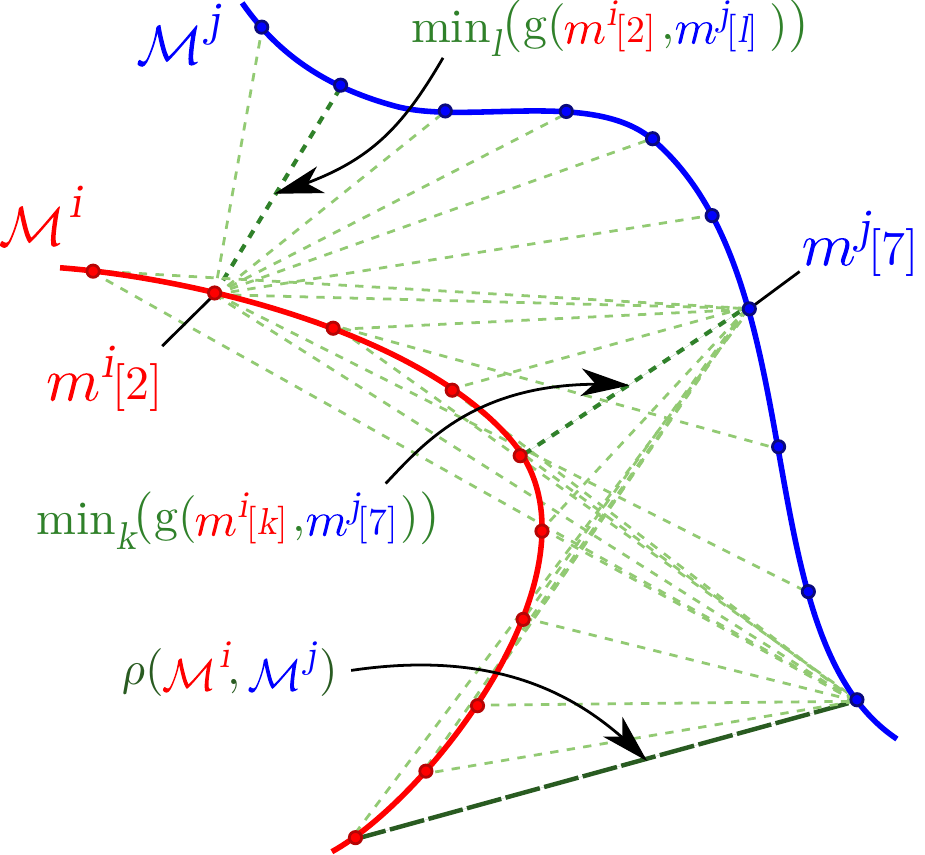}
\caption{The Hausdorff distance between two sets $\mathcal{M}^i$ and $\mathcal{M}^j$ corresponds to the greatest of all the distances from a point in one set to the closest point in the other set.}
\label{fig:Hausdorff}
\end{figure}
The Hausdorff distance between two sets is the greatest of the shortest distances taken to travel from one set to the other in the metric space they belong to (see illustration in Fig.\ref{fig:Hausdorff}). In our application, the metric space is Euclidean (motor space) and the two sets are the sampled manifolds $\mathcal{M}^i$ and $\mathcal{M}^j$. However, we slightly modify the Euclidean metrics to take into account the periodicity implied by the agent's hinge degrees of freedom. Indeed, the external retinal configuration, and thus the sensory input $\mathbf{s}$, is unchanged modulo $2\pi$ for any motor output $m_i, \forall i \in \{1,\dots,4\}$. This periodicity would artificially increase the Euclidean distance between motor outputs which should be closer according to sensory information (internal point of view) and to the real configuration of the robot (external point of view).
We thus use a modified Euclidean distance such that the distance between the $k$-th sample of $\mathcal{M}^i$ and the $l$-th sample of $\mathcal{M}^j$ in the motor space is computed as follows:
\begin{equation}
	g\big(\mathbf{m}^i[k],\mathbf{m}^j[l]\big)=  \sqrt{\sum_{n=1}^{N} h\Big(m^i_n[k]-m^j_n[l] \Big)^2},
\label{eq:deuclimodif}
\end{equation}
with $N=4$ the dimension of the motor space and:
\begin{equation}	
h(u) = \left\{
\begin{array}{r c c}
	2\pi-u  & \text{ if } & u>\pi \\
	-2\pi-u & \text{ if } & u<-\pi
\end{array}
\right\}.
\end{equation}
This way, the distance between two individual motor values $\{m^i_n[k],m^j_n[l]\}$ cannot be greater than $\pi$, while the distance between two motor configurations $\{\mathbf{m}^i[k],\mathbf{m}^j[l]\}$ cannot be greater than $\sqrt{4\pi^2}=2\pi$.

Finally, the modified Hausdorff distance between $\mathcal{M}^i$ and $\mathcal{M}^j$ is computed as follows:
\begin{equation}
\rho(\mathcal{M}^i,\mathcal{M}^j) = \max \left\{
	\begin{array}{c}
		\max_k \Big( \min_l \Big( g\big(\mathbf{m}^i[k],\mathbf{m}^j[l]\big) \Big) \Big) \\
		\max_l \Big( \min_k \Big( g\big(\mathbf{m}^i[k],\mathbf{m}^j[l]\big) \Big) \Big) \\
	\end{array}
\right\}.
\end{equation}

The definition of all pairwise distances allows us to project the underlying manifold of $\mathcal{M}$'s. The latter is embedded into a space of a priori unknown dimensionality, but here it is projected into a $3$D space for the visualization purpose.
It is important to notice here again that this projection isn't mandatory and doesn't have to be explicitly performed by the agent. Indeed, all relevant information about the manifold of points of view is already contained in the metrics $\rho(\mathcal{M}^i,\mathcal{M}^j)$. By simply applying this new metrics instead of the natural Euclidean one in the motor space, the agent can for example reduce the intrinsic dimensionality of its motor sampling when exploring a new environment. Hereunder the projection is nonetheless a useful tool to visualize and analyze the manifold structure captured by the agent.
It was performed using a non-linear dimension reduction method: Curvilinear Component Analysis (CCA) \cite{Demartines1997}. The CCA defines a position in the representational space for every projected point by trying to preserve the topology of the underlying manifold. For more details on the method, a complete description of the algorithm can be found in~\cite{Demartines1997} and an example application in~\cite{Laflaquiere2012}.

\subsection{Results and discussion}
\label{sec:Part3-3}
\begin{figure}[h!]
\centering
\includegraphics[height=0.83\textheight]{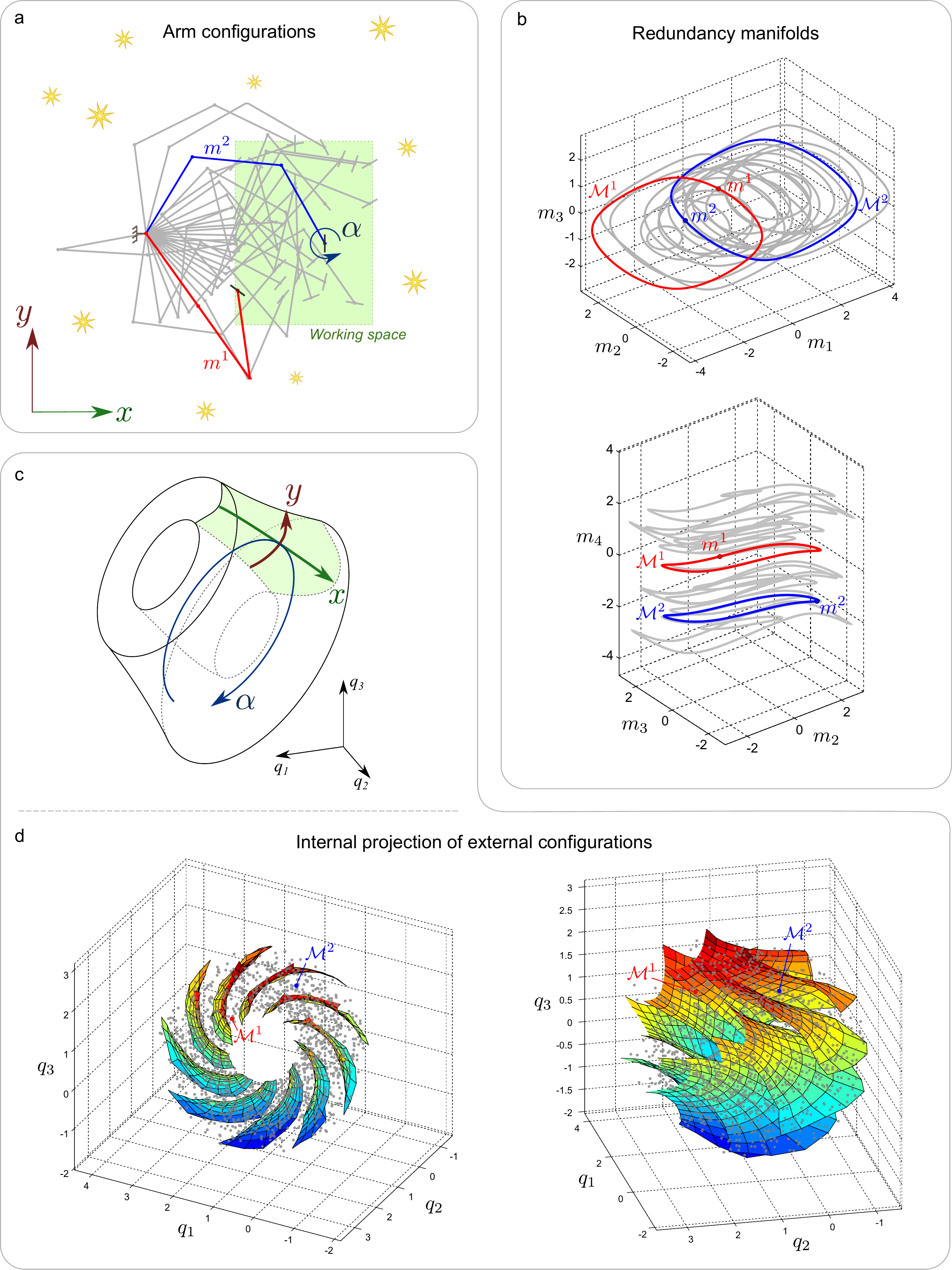}
\caption{a) $1$\% of the $2500$ exploratory arm configurations $\mathbf{m}^i$. b) Two $3$D projections of $1$\% of the sets $\mathcal{M}^i$ embedded in the $4$D motor space. c) Schematic of the projected manifold and capturing of external parameters. d) Projection in $3$D of the $2500$ manifolds $\mathcal{M}^i$ (gray points) with surfaces corresponding to translations in the working space for different retinal orientations.}
\label{fig:Arm_manifold}
\end{figure}

The resulting projection is provided in Fig.~\ref{fig:Arm_manifold}d, with two different views on the manifold. Colored surfaces have been added to facilitate its interpretation. Each one corresponds to a $10\times 10$ uniform sampling of positions $[x,y]$ in the working space for a fixed orientation $\alpha$. Depending on the surface, this orientation varies from $0$ to $360$ degrees in $36$-degrees steps. The color coding of the surfaces is associated, from an external point of view, to the parameter $x$.
Figure \ref{fig:Arm_manifold}c presents a qualitative representation of the manifold; a section associated to the projection of the working space for a given orientation of the retina is colored in green.
As expected for the configuration of a planar object in plane, the resultant manifold has the topology of $\mathbb T\times\mathbb{R}^2$; that is, a circle $\mathbb T$ (retinal orientation) times a plane $\mathbb{R}^2$ (retinal position).
As displayed schematically in Fig.\ref{fig:Arm_manifold}c, each parameter of the retina's external configuration $[x,y,\alpha]$ has been successfully captured: $x$ is encoded along the manifold's height, $y$ along its width and $\alpha$ along its transverse section.
By discovering sets $\mathcal{M}^i$ associated with constant sensory inputs and determining the structure of the manifold they form, the agent has thus been able to build an internal representation of its external configuration. Moreover, this manifold is built from motor sets $\mathcal{M}^i$ which are environment-independent. Performing the same exploration in different environments thus leads to the construction of the same internal manifold. As described in~\S\ref{sec:Part1_2}, some peculiar environments could lead to the definition of different sets $\mathcal{M}^i$ even if the structure of the agent has not changed. If they are rare, these exotic cases would be statistically filtered out over the agent's lifetime experience. If they are recurrent, a different structure would be captured by the agent, leading to an altered notion of space, but it would perfectly characterize the agent's interaction with the world and the actual spatial experience that it can have. Nonetheless, in rich enough environments the probability of having such symmetric sensory experiences should be insignificant.

It is important to distinguish the agent's internal representation of space and the illustration of this representation, depicted in Fig.\ref{fig:Arm_manifold}. The internal representation is captured by the sets $\mathcal{M}^i$ and the manifold of these sets obtained by introducing a metric on them (which is the Hausdorff distance in our example). The illustration of this representation is obtained by immersing the manifold of $\mathcal M$'s into a vector space and then showing its non-linear projection (CCA) into a three-dimensional space (see Fig.\ref{fig:Arm_manifold}). The purpose of the illustration is exclusively to facilitate the understanding of the internal representation for a reader. The illustration is rather unstable. For example, for a different working space (shape, size or position), the final projection would change. As CCA cannot preserve the metric precisely during data projection, the orientation, position and to some extent the shape of the projection would be (randomly) different from the one displayed in Fig.\ref{fig:Arm_manifold}d. Nevertheless, the topology of the two manifolds would be coherent. In other words, the manifolds generated by the separate exploration of two different working spaces could be patched together so as to appear as two connected pieces of a single larger manifold. This consistency highlights the fact that the agent is exploring a single external space, possibly in separate patches. Likewise, the structure of the constraints imposed by space on the agent's sensorimotor experience could be captured incrementally by progressively discovering the sets $\mathcal{M}^i$. When such a new experience is available, the distances to the previously stored sets have to be estimated to extend the internal manifold representing the agent's external configuration.

Furthermore, it is important to remark that the internal manifold generated in the last experiment can only be visualized because the size of the agent's working space is limited. It should not be possible to represent a manifold associated with an object configuration on the plane (two translational coordinates and orientation angle) in $3$D because of the circularity involved in orientation. The use of a limited working space allows the manifold to be strongly deformed during the projection in $3$D to take the circular dimension into account while avoiding any self-collision. For wider working spaces, the manifold could not be visualized in $3$D but would still be explicitly described by the metrics defined on the sets $\mathcal{M}^i$. It would only be possible to visualize it locally.

Importantly, note that the technical implementation used here (working space exploration, sampling of redundant manifolds, metric computation, low-dimensional projection) is not the main focus of the paper. The scope of the approach is generic, and it can be extended to different implementations of each step of data processing step (such as using ISOMAP \cite{Tenenbaum2000} instead of CCA, or using a different metric on the manifolds $\mathcal{M}^i$) and, of course, to other agent/environment systems. Applying the method to more complex systems would, however, lead to some technical challenges that we have not addressed in this paper. The main difficulty, as pointed out in many studies in developmental robotics~\cite{Baranes:2013}, would be the exploration of the motor space. Random exploration becomes inefficient as dimensionality increases. Some heuristics or bootstrapping reflexes thus need to be introduced to guide the sampling of the motor space \cite{oudeyer2007}. For instance, a low-level tracking-like behavior could facilitate the sampling of the sets $\mathcal{M}^i$, whose dimensionality would be greater than $1$ for more complex robots. Note, however, that learning space is a challenge for humans as well. Spatial reasoning remains immature until the age of 6-10 years~\cite{Dillon2013}. Evidently, so many years of sensorimotor experience provide an enormous amount of data to lift the curse of dimensionality. Having a robot explore its environment for an extended period of time may thus be an inevitable cost to be paid for the autonomous emergence of such perceptual notions.

Finally, the results highlight the fact that the proposed approach relies entirely on the properties of the raw sensorimotor flow. In contrast, as mentioned in the introduction, most work on body schema acquisition hypothesizes some spatial knowledge provided a priori to the robot, through the pre-processing of sensory inputs or through additional knowledge on the agent’s own structure~\cite{Bongard2006}. The only assumption in the present work is the hypothesis that the metric on the sets $\mathcal{M}^i$ is topologically identical to the metric of the agent's external configuration. We thus believe that any perceptive knowledge should be discovered and grounded in sensorimotor experience.
It should also be noted that some papers offer an implicit definition of body schema based on learning direct relations between motor commands and sensory inputs~\cite{Natale:2007, Fuke:2007}. This behavioralist approach suggests that the notion of space is somehow irrelevant to having a robot act in the world. Although this may be true for simple behaviors, we think that a compact internal representation of the world is required to create new complex behaviors. In particular, space is such a ubiquitous component of our perception that it must correspond to fundamental properties of our sensorimotor experience. Capturing them enables the efficient interpretation of future sensory information.

\section{Conclusion}
\label{sec:Discussion}

The objective of this paper was to understand how a naive \emph{tabula rasa} agent can capture properties of the external space it is immersed in. The question, more precisely, is: how can it discover that such a structure exists independently of the particular objects contained in the environment when its sensory experience is heavily environment-dependent?
Although the notion of space is intuitively associated with visual inputs (or more generally with exteroception), we show that considering them exclusively cannot lead to an environment-independent notion. We thus proposed to look for spatial properties not in the agent's exteroceptive flow but in the structure of its sensorimotor interactions with the world. From the agent's point of view, the sensorimotor laws are just functions mapping its motor outputs to sensory inputs, and both the properties of the external world and its own properties are simply constraints on the shape of these functions. In particular, the structure of the external space as such manifests itself through the constraint that certain properties of these functions should not depend on the objects present in the environment. We suggested that the discovery of such constraints can lead the agent to discover the structure of external space.

The method we developed allows the agent, starting with minimal a priori knowledge, to capture those constraints and to build an internal representation of its external configuration in space. It requires the agent to notice that different motor commands can result in identical sensory inputs. This discovery reveals the existence of an underlying structure of motor sets ($\mathcal{M}^i$) in the motor space. This structure can be processed as a manifold by defining a metric on the motor sets.  Interestingly, from an external point of view, this manifold corresponds to the manifold of external agent's configurations in space. It is important to note, though, that to perform such an analysis, the agent does not need to assume the existence of space. Its final internal representation can simply be the result of an effort to generate a compact and invariant encoding of its sensorimotor interactions with the world.

Even if space was the main focus of this work, the processing and results presented in this paper can be linked to the rich literature on body schema acquisition and forward model learning. Indeed, the notion of space is intimately related to the agent's ability to move \cite{Poincare1895}. However, our approach differs from a large part of this literature by not assuming the pre-coding of building blocks or any processing of the sensory flow. In the line of this paper, other works focus directly on the raw sensorimotor flow to build a body schema and/or control a robot \cite{Censi2012,Hoffmann2013,Kuipers2008,Rolf2014}. However these approaches rarely address the question of space, instead focusing on the emergence of behaviors (however, see \cite{Roschin2011}). Our work adds to existing research by making explicit what properties of the raw sensorimotor flow can lead an agent to perceive space.

In this paper, we have focused on capturing one property of space: namely, the fact that the notion of space is independent of the content of the environment. However, this single property does not capture the whole concept of space as we mean it from our subjective point of view. As also pinpointed by Poincar\'e~\cite{Poincare1895}, space is felt as \emph{a container shared by both the agent and the environment}. We have the subjective experience that we and the objects around us are immersed in the same space. Our next goal is to ground this property in the agent's sensorimotor experience. To do so, we will focus on discovering the displacements that space allows both the agent and objects around it to perform. Capturing such specific experiences will also overcome a limitation of the present work: the structure discovered in this paper is an internal representation of the agent's external configuration, but this broad notion includes both spatial and non-spatial parameters. One can, for example, imagine an agent similar to the one described in~\S\ref{sec:Part3} but with an additional form of motor control linked to a pupil that adjusts the amount of light arriving on the retina. Such a non-spatial degree of freedom would be captured without distinction by our method, increasing the dimensionality of the internal manifold. By searching only for displacements or, in sensorimotor terms, transformations that both the agent and objects in the environment can perform, it should be possible to avoid capturing such non-spatial degrees of freedom. The final structure captured by the agent will thus be closer to what we intuitively expect from a notion of space.

\paragraph{Acknowledgment} The authors would like to thank anonymous reviewers for their useful comments contributing towards better positioning and clarity of the manuscript. The authors would also like to thank Paul Reeve for his editorial corrections.

\clearpage


\appendix

\section{Sampling of the redundant manifolds $\mathcal{M}^i$}
\label{sec:Appendix A}
Let $\mathbf{m}^i=\mathbf{m}^i[1]$ be an initial motor configuration generated during the exploration. The manifold $\mathcal{M}^i$ it belongs to is sampled using the kernel of the Jacobian $J$ of the arm at $\mathbf{m}^i[1]$:
\begin{equation}
J  =  \left(
	\begin{array}{c c c c}
	\frac{\partial \alpha}{\partial m_1} & \frac{\partial \alpha}{\partial m_2} & \frac{\partial \alpha}		{\partial m_3} & \frac{\partial \alpha}{\partial m_4}\\
	\frac{\partial x}{\partial m_1} & \frac{\partial x}{\partial m_2} & \frac{\partial x}{\partial m_3} & \frac	{\partial x}{\partial m_4}\\
	\frac{\partial y}{\partial m_1} & \frac{\partial y}{\partial m_2} & \frac{\partial y}{\partial m_3} & \frac	{\partial y}{\partial m_4}
	\end{array}
	\right) \Bigg \rfloor_{\mathbf{m}^i[1]} \nonumber
\end{equation}
\begin{equation}
  = \left(
	\begin{array}{c c c c}
	1 & 1 & 1 & 1\\
	-\sin_{1}-\sin_{12}-\sin_{123} & -\sin_{12}-\sin_{123} & -\sin_{123} & 0\\
	\cos_{1}+\cos_{12}+\cos_{123} & \cos_{12}+\cos_{123} & \cos_{123} & 0\\
	\end{array}
	\right )
\end{equation}
with $\sin_{1}=\sin(m^i_1[1])$, $\sin_{12}=\sin(m^i_1[1]+m^i_2[1])$, $\sin_{123}=\sin(m^i_1[1]+m^i_2[1]+m^i_3[1])$, $\cos_{1}=\cos(m^i_1[1])$, $\cos_{12}=\cos(m^i_1[1]+m^i_2[1])$ and $\cos_{123}=\cos(m^i_1[1]+m^i_2[1]+m^i_3[1])$.

In the motor space, the kernel of $J$ corresponds to a vector $\mathbf{v}[1]$ along which any local motor change does not produce any movement of the arm's end-point. It is estimated by performing a singular value decomposition (SVD) of $J$ and retaining the right singular vector associated with a null singular value. A new redundant motor configuration $\mathbf{m}^i[2]$ can be generated by starting from $\mathbf{m}^i[1]$ and moving a small distance along $\mathbf{v}[1]$. The same process can be generalized to iteratively generate new samples:
\begin{equation}
	\mathbf{m}^i[j+1] = \mathbf{m}^i[j] + \mu \rho \mathbf{v}[j],
\end{equation}
with $\mu = 10^{-3}$ a small step length, $\mathbf{v}[j]$ the kernel of $J$ at $\mathbf{m}^i[j]$ and $\rho$ a factor ensuring that the manifold is constantly sampled in the same  direction:
\begin{equation}
\label{eq:rho}
\rho = \left\{
	\begin{array}{c c c}
		1  & \text{ if } & \mathbf{v}[j] \cdot \mathbf{v}[j-1] \geq 0\\
		-1 & \text{ if } & \mathbf{v}[j] \cdot \mathbf{v}[j-1] < 0 \\
		1  & \text{ if } & j=0
	\end{array}
	\right\}.
\end{equation}
The rotary nature of the arm's actuators implies that the manifold $\mathcal{M}^i$ is closed. This is true unless the distance from the retina to the agent's base is less than the length of one arm segment, which is not the case in our working space. (see Fig.\ref{fig:Arm_manifold}).
If this were the case, the manifold $\mathcal{M}^i$ would be split into two disjoint submanifolds due to the arm's mechanical structure. The proposed method based on the use of $J$ to explore $\mathcal{M}^i$ would then be unsuitable. Note that another exploration method, such as an exhaustive exploration of the motor space, could overcome this limitation.

New samples are generated until the following condition is fulfilled:
\begin{equation}
	||\mathbf{m}^i[j],\mathbf{m}^i[1]|| \leq 10^{-2}, \text{ with } j \geq 50.
\end{equation}
Finally, the number of samples $\mathbf{m}^i[j]$ can be different for different manifolds $\mathcal{M}^i$. It is thus homogenized through a standard interpolation in order to retain only $100$ regularly distributed samples for each manifold.
\clearpage

\section{Algorithm}
\label{sec:Algorithm}
    \begin{algorithm}
      {{M}}: number of configurations generated in the working space during exploration.\\
      {{conf\_space}}: joints limits defining the working space.\\
      {{$J$}}: Jacobian of the robot.\\
      {{$\epsilon$}}: threshold detecting the loop closing of the motor kernel manifold.\\
      {{$\mu$}}: step size for the sampling of the motor kernel manifold (see~\ref{sec:Appendix A}).
      \caption{Generation of the internal representation}
{\footnotesize{
        \begin{algorithmic}[1]
          \REQUIRE M, conf\_space, $J$, $\epsilon$, $\mu$
          \STATE $i$=1;
          \WHILE{$i\leq M$}
             \STATE $j$=1;
             \STATE
             \STATE \COMMENT{Pick a random motor configuration}
             \STATE $\mathbf{m}^i[j] = (m_1^i, m_2^i, m_3^i, m_4^i)^T = \text{rand}(4,1)$; 
             \IF{$\mathbf{m}^i[j] \notin$ conf\_space} \RETURN{Line 3; \emph{\% The end effector is out of the working space}} \ENDIF
             \STATE
             \STATE \COMMENT{Sampling of the corresponding kernel manifold}
             \REPEAT 
                 \STATE{$\mathbf{v}[j] = \left.\ker J\right\vert_{{\mathbf{m}}_i [j]}$;} 
                 \STATE{Compute $\rho$ according to \eqref{eq:rho}}
                 \STATE{$\mathbf{m}^i[j+1] = \mathbf{m}^i[j] + \mu\rho \mathbf{v}[j]$;} 
                 \STATE{$j=j+1$;}
             \UNTIL{$\|\mathbf{m}^i[j+1] - \mathbf{m}^i[1]\| < \epsilon$ AND $j>50$;}
             \STATE $\mathcal{M}^i = (\mathbf{m}^i[1], \ldots, \mathbf{m}^i[j])^T$;
             \STATE
             \STATE $i=i+1$;
           \ENDWHILE
           \STATE
           \STATE \COMMENT{Computation of the metric}
           \FOR{i=1:M} 
               \FOR{k=1:M} 
                    \STATE{$X(i,k) = \rho(\mathcal{M}^i, \mathcal{M}^k)$;} 
               \ENDFOR
           \ENDFOR
          \STATE
           \STATE \COMMENT{Computation of a low-dimensional projection through CCA}
           \STATE $C$ = cca($X$);
          \RETURN $C$
        \end{algorithmic}
}}
    \end{algorithm}

\clearpage

\bibliography{biblio-alban}

\begin{thebibliography}{10}
\expandafter\ifx\csname url\endcsname\relax
  \def\url#1{\texttt{#1}}\fi
\expandafter\ifx\csname urlprefix\endcsname\relax\def\urlprefix{URL }\fi
\expandafter\ifx\csname href\endcsname\relax
  \def\href#1#2{#2} \def\path#1{#1}\fi

\bibitem{Sigaud2011}
O.~Sigaud, C.~Sala\"{u}n, V.~Padois, {On-line regression algorithms for
  learning mechanical models of robots: a survey}, Robotics and Autonomous
  Systems 59~(12) (2011) 1115--1129.

\bibitem{Hersch2008}
M.~Hersch, E.~L. Sauser, A.~Billard, {Online Learning of the Body Schema}, I.
  J. Humanoid Robotics 5~(2) (2008) 161--181.

\bibitem{Sturm2009}
J.~Sturm, C.~Plagemann, W.~Burgard, Body schema learning for robotic
  manipulators from visual self-perception, Journal of Physiology-Paris
  103~(3–5) (2009) 220 -- 231, neurorobotics.
\newblock \href
  {http://dx.doi.org/http://dx.doi.org/10.1016/j.jphysparis.2009.08.005}
  {\path{doi:http://dx.doi.org/10.1016/j.jphysparis.2009.08.005}}.

\bibitem{Yoshikawa2004}
Y.~Yoshikawa, Y.~Tsuji, K.~Hosoda, M.~Asada, Is it my body? body extraction
  from uninterpreted sensory data based on the invariance of multiple sensory
  attributes, in: Intelligent Robots and Systems, 2004. (IROS 2004).
  Proceedings. 2004 IEEE/RSJ International Conference on, Vol.~3, 2004, pp.
  2325--2330 vol.3.
\newblock \href {http://dx.doi.org/10.1109/IROS.2004.1389756}
  {\path{doi:10.1109/IROS.2004.1389756}}.

\bibitem{Martinez2010}
R.~Martinez-Cantin, M.~Lopes, L.~Montesano, Body schema acquisition through
  active learning, in: Robotics and Automation (ICRA), 2010 IEEE International
  Conference on, 2010, pp. 1860--1866.
\newblock \href {http://dx.doi.org/10.1109/ROBOT.2010.5509406}
  {\path{doi:10.1109/ROBOT.2010.5509406}}.

\bibitem{Hoffmann2010}
M.~Hoffmann, H.~G. Marques, A.~{Hernandez Arieta}, H.~Sumioka, M.~Lungarella,
  R.~Pfeifer, {Body Schema in Robotics: A Review}, IEEE Transactions on
  Autonomous Mental Development 2~(4) (2010) 304--324.
\newblock \href {http://dx.doi.org/10.1109/TAMD.2010.2086454}
  {\path{doi:10.1109/TAMD.2010.2086454}}.

\bibitem{Bongard2006}
J.~Bongard, V.~Zykov, H.~Lipson, Resilient machines through continuous
  self-modeling., Science 314~(5802) (2006) 1118--1121.
\newblock \href {http://dx.doi.org/10.1126/science.1133687}
  {\path{doi:10.1126/science.1133687}}.

\bibitem{Koos2013}
S.~Koos, A.~Cully, J.-B. Mouret, Fast damage recovery in robotics with the
  t-resilience algorithm, The International Journal of Robotics Research
  32~(14) (2013) 1700--1723.

\bibitem{Wiesel2005}
F.~Wiesel, O.~Tenchio, M.~Simon, et~al., Reinforcing the driving quality of
  soccer playing robots by anticipation, it--Information Technology 47 (2005)
  5.

\bibitem{Tsai1988}
R.~Y. Tsai, R.~K. Lenz, Real time versatile robotics hand/eye calibration using
  3d machine vision, in: Robotics and Automation, 1988. Proceedings., 1988 IEEE
  International Conference on, IEEE, 1988, pp. 554--561.

\bibitem{Bennett1991}
D.~J. Bennett, D.~Geiger, J.~M. Hollerbach, Autonomous robot calibration for
  hand-eye coordination, The International journal of robotics research 10~(5)
  (1991) 550--559.

\bibitem{Schilling2011}
M.~Schilling, Universally manipulable body models—dual quaternion
  representations in layered and dynamic mmcs, Autonomous Robots 30~(4) (2011)
  399--425.

\bibitem{Hersch2009}
M.~Hersch, Adaptive sensorimotor peripersonal space representation and motor
  learning for a humanoid robot, Ph.D. thesis, Citeseer (2009).

\bibitem{Chao2014}
F.~Chao, M.~H. Lee, M.~Jiang, C.~Zhou, {An Infant Development-inspired Approach
  to Robot Hand-eye Coordination}, Int J Adv Robot Syst 11:15 (2014) 1--14.

\bibitem{Jamone2012}
L.~Jamone, L.~Natale, G.~Metta, F.~Nori, G.~Sandini, {Autonomous Online
  Learning of Reaching Behavior in a Humanoid Robot}, International Journal of
  Humanoid Robotics 9~(3).

\bibitem{NataleMetta2007}
L.~Natale, F.~Nori, G.~Sandini, G.~Metta, Learning precise 3d reaching in a
  humanoid robot, in: Development and Learning, 2007. ICDL 2007. IEEE 6th
  International Conference on, IEEE, 2007, pp. 324--329.

\bibitem{Gaskett2003}
C.~Gaskett, G.~Cheng, Online learning of a motor map for humanoid robot
  reaching.

\bibitem{Rensink1997}
R.~A. Rensink, J.~K. O'Regan, J.~J. Clark, To see or not to see: The need for
  attention to perceive changes in scenes, Psychological science 8~(5) (1997)
  368--373.

\bibitem{ORegan1999}
J.~K. O'Regan, R.~A. Rensink, J.~J. Clark, {Change-blindness as a result of
  ‘mudsplashes’}, Nature 398~(6722) (1999) 34.

\bibitem{Benosman2011}
R.~Benosman, S.-H. Ieng, P.~Rogister, C.~Posch, Asynchronous event-based
  hebbian epipolar geometry, Neural Networks, IEEE Transactions on 22~(11)
  (2011) 1723--1734.

\bibitem{Lorach2012}
H.~Lorach, R.~Benosman, O.~Marre, S.-H. Ieng, J.~A. Sahel, S.~Picaud,
  Artificial retina: the multichannel processing of the mammalian retina
  achieved with a neuromorphic asynchronous light acquisition device, Journal
  of neural engineering 9~(6) (2012) 066004.

\bibitem{Rogister2012}
P.~Rogister, R.~Benosman, S.-H. Ieng, P.~Lichtsteiner, T.~Delbruck,
  Asynchronous event-based binocular stereo matching, Neural Networks and
  Learning Systems, IEEE Transactions on 23~(2) (2012) 347--353.

\bibitem{Censi2012}
A.~Censi, R.~M. Murray, Learning diffeomorphism models of robotic sensorimotor
  cascades, in: Robotics and Automation (ICRA), 2012 IEEE International
  Conference on, IEEE, 2012, pp. 3657--3664.

\bibitem{Hoffmann2013}
N.~M. Schmidt, M.~Hoffmann, K.~Nakajima, R.~Pfeifer, Bootstrapping perception
  using information theory: Case studies in a quadruped robot running on
  different grounds, Advances in Complex Systems 16~(02n03).

\bibitem{Kuipers2008}
B.~Kuipers, Drinking from the firehose of experience, Artificial Intelligence
  in Medicine 44~(2) (2008) 155--170.

\bibitem{Schmidhuber1996}
J.~Schmidhuber, M.~Eldracher, B.~Foltin, {Semilinear Predictability
  Minimization Produces Well-Known Feature Detectors}, Neural Computation 8
  (1996) 773--786.

\bibitem{Olshausen1996}
B.~A. Olshausen, D.~J. Field, {Emergence of simple-cell receptive field
  properties by learning a sparse code for natural images.}, Nature 381~(6583)
  (1996) 607--609.
\newblock \href {http://dx.doi.org/10.1038/381607a0}
  {\path{doi:10.1038/381607a0}}.

\bibitem{Masquelier2007}
T.~Masquelier, S.~J. Thorpe, {Unsupervised learning of visual features through
  spike timing dependent plasticity}, PLoS Computational Biology 3~(2) (2007)
  e31.

\bibitem{Lee2008}
H.~Lee, C.~Ekanadham, A.~Ng, {Sparse deep belief net model for visual area V2},
  in: J.~C. Platt, D.~Koller, Y.~Singer, S.~Roweis (Eds.), Advances in Neural
  Information Processing Systems 20, MIT Press, Cambridge, MA, 2008, pp.
  873--880.

\bibitem{Choe2008}
Y.~Choe, H.-F. Yang, N.~Misra, Motor system’s role in grounding, receptive
  field development, and shape recognition, 2008 7th IEEE International
  Conference on Development and Learning 1 (2008) 67--72.
\newblock \href {http://dx.doi.org/10.1109/DEVLRN.2008.4640807}
  {\path{doi:10.1109/DEVLRN.2008.4640807}}.

\bibitem{Klyubin2004}
E.~S. Klyubin, D.~Polani, C.~L. Nehaniv, {Tracking Information Flow through the
  Environment: Simple Cases of Stigmergy}, in: Artificial Life IX: Proceedings
  of the Ninth International Conference on the Simulation and Synthesis of
  Living Systems, The MIT Press, 2004, pp. 563--568.

\bibitem{McGregor2011}
S.~McGregor, D.~Polani, K.~Dautenhahn, {Generation of tactile maps for
  artificial skin}, PloS one 6~(11) (2011) e26561.

\bibitem{Kaplan2004}
F.~Kaplan, P.-Y. Oudeyer, Maximizing learning progress: an internal reward
  system for development, in: Embodied artificial intelligence, Springer, 2004,
  pp. 259--270.

\bibitem{Klyubin2005a}
A.~S. Klyubin, D.~Polani, C.~L. Nehaniv, Empowerment: A universal agent-centric
  measure of control, in: Evolutionary Computation, 2005. The 2005 IEEE
  Congress on, Vol.~1, IEEE, 2005, pp. 128--135.

\bibitem{Gordon2011}
G.~Gordon, E.~Ahissar, Reinforcement active learning hierarchical loops, in:
  Neural Networks (IJCNN), The 2011 International Joint Conference on, IEEE,
  2011, pp. 3008--3015.

\bibitem{OReganNoe2001}
J.~K. O'Regan, A.~No\"{e}, {A sensorimotor account of vision and visual
  consciousness.}, The Behavioral and brain sciences 24~(5) (2001) 939--73;
  discussion 973--1031.

\bibitem{Pierce1997}
D.~Pierce, B.~J. Kuipers, {Map learning with uninterpreted sensors and
  effectors}, Artificial Intelligence 92~(1) (1997) 169--227.

\bibitem{Mueller1988}
M.~Müller, R.~Wehner, Path integration in desert ants, cataglyphis fortis.,
  Proc Natl Acad Sci U S A 85~(14) (1988) 5287--5290.

\bibitem{Smith1990}
R.~Smith, M.~Self, P.~Cheeseman,
  \href{http://dx.doi.org/10.1007/978-1-4613-8997-2_14}{Estimating uncertain
  spatial relationships in robotics}, in: I.~Cox, G.~Wilfong (Eds.), Autonomous
  Robot Vehicles, Springer New York, 1990, pp. 167--193.
\newblock \href {http://dx.doi.org/10.1007/978-1-4613-8997-2_14}
  {\path{doi:10.1007/978-1-4613-8997-2_14}}.
\newline\urlprefix\url{http://dx.doi.org/10.1007/978-1-4613-8997-2_14}

\bibitem{Bowling2007}
M.~Bowling, D.~Wilkinson, A.~Ghodsi, A.~Milstein, Subjective localization with
  action respecting embedding, in: S.~Thrun, R.~Brooks, H.~Durrant-Whyte
  (Eds.), Robotics Research, Vol.~28 of Springer Tracts in Advanced Robotics,
  Springer Berlin Heidelberg, 2007, pp. 190--202.
\newblock \href {http://dx.doi.org/10.1007/978-3-540-48113-3_18}
  {\path{doi:10.1007/978-3-540-48113-3_18}}.

\bibitem{Poincare1895}
H.~Poincar\'{e}, {L'espace Et la G\'{e}om\'{e}trie}, Revue de M\'{e}taphysique
  Et de Morale 3~(6) (1895) 631--646.

\bibitem{Stober2011}
J.~Stober, R.~Miikkulainen, B.~Kuipers, {Learning geometry from sensorimotor
  experience}, in: Development and Learning (ICDL), 2011 IEEE International
  Conference on, Vol.~2, IEEE, 2011, pp. 1--6.

\bibitem{Roschin2011}
V.~Y. Roschin, A.~A. Frolov, Y.~Burnod, M.~A. Maier, {A neural network model
  for the acquisition of a spatial body scheme through sensorimotor
  interaction}, Neural computation 23~(7) (2011) 1821--1834.

\bibitem{Philipona2003}
D.~Philipona, J.~K. O'Regan, J.-P. Nadal, {Is there something out there?:
  Inferring space from sensorimotor dependencies}, Neural Comput. 15~(9) (2003)
  2029--2049.
\newblock \href {http://dx.doi.org/10.1162/089976603322297278}
  {\path{doi:10.1162/089976603322297278}}.

\bibitem{Laflaquiere2012}
A.~Laflaquiere, S.~Argentieri, O.~Breysse, S.~Genet, B.~Gas, {A non-linear
  approach to space dimension perception by a naive agent}, in: Intelligent
  Robots and Systems (IROS), 2012 IEEE/RSJ International Conference on, IEEE,
  2012, pp. 3253--3259.

\bibitem{Laflaquiere2010}
A.~Laflaquiere, S.~Argentieri, B.~Gas, E.~Castillo-Castenada, {Space dimension
  perception from the multimodal sensorimotor flow of a naive robotic agent},
  in: Intelligent Robots and Systems (IROS), 2010 IEEE/RSJ International
  Conference on, IEEE, 2010, pp. 1520--1525.

\bibitem{Bernard2012}
M.~Bernard, P.~Pirim, A.~de~Cheveign�, B.~Gas, {Sensorimotor Learning of
  Sound Localization from an Auditory Evoked Behavior}, in: IEEE International
  Conference on Robotics and Automation, ICRA2012, St. Paul, MN, USA, 2012, pp.
  91--96.

\bibitem{Laflaquiere2013}
A.~Laflaquiere, A.~V. Terekhov, B.~Gas, J.~K. O'Regan, {Learning an internal
  representation of the end-effector configuration space}, in: Intelligent
  Robots and Systems (IROS), 2013 IEEE/RSJ International Conference on, IEEE,
  2013, pp. 1230--1235.

\bibitem{Terekhov2013}
A.~V. Terekhov, J.~K. O'Regan, {Space as an invention of biological organisms},
  arXiv preprint arXiv:1308.2124.

\bibitem{Wyss2006}
R.~Wyss, P.~K{\"{o}}nig, P.~F. M.~J. Verschure,
  \href{http://dx.doi.org/10.1371/journal.pbio.0040120}{A model of the ventral
  visual system based on temporal stability and local memory.}, PLoS Biol 4~(5)
  (2006) e120.
\newblock \href {http://dx.doi.org/10.1371/journal.pbio.0040120}
  {\path{doi:10.1371/journal.pbio.0040120}}.
\newline\urlprefix\url{http://dx.doi.org/10.1371/journal.pbio.0040120}

\bibitem{Cheah2010}
C.~Cheah, X.~Li, Reach then see: A new adaptive controller for robot
  manipulator based on dual task-space information, in: Robotics and Automation
  (ICRA), 2010 IEEE International Conference on, 2010, pp. 5155--5160.
\newblock \href {http://dx.doi.org/10.1109/ROBOT.2010.5509956}
  {\path{doi:10.1109/ROBOT.2010.5509956}}.

\bibitem{Held1963}
R.~Held, A.~Hein, {Movement-produced stimulation in the development of visually
  guided behavior}, J Comp Physiol Psychol (1963) 872--876.

\bibitem{Bach-y-Rita:2003it}
P.~B.-y. Rita, S.~W. Kercel, Sensory substitution and the human-machine
  interface, Trends in Cognitive Sciences 7~(12) (2003) 541--546.

\bibitem{Blakemore1970}
C.~Blakemore, G.~F. Cooper, Development of the brain depends on the visual
  environment., Nature 228~(5270) (1970) 477--478.

\bibitem{Demartines1997}
P.~Demartines, J.~Herault, {Curvilinear component analysis: a self-organizing
  neural network for nonlinear mapping of data sets.}, IEEE transactions on
  neural networks / a publication of the IEEE Neural Networks Council 8~(1)
  (1997) 148--54.
\newblock \href {http://dx.doi.org/10.1109/72.554199}
  {\path{doi:10.1109/72.554199}}.

\bibitem{Tenenbaum2000}
J.~Tenenbaum, V.~de~Silva, J.~Langford, {A Global Geometric Framework for
  Nonlinear Dimensionality Reduction}, Science 290~(5500) (2000) 2319--2323.
\newblock \href {http://dx.doi.org/doi: 10.1126/science.290.5500.2319}
  {\path{doi:doi: 10.1126/science.290.5500.2319}}.

\bibitem{Baranes:2013}
A.~Baranes, P.-Y. Oudeyer, Active learning of inverse models with intrinsically
  motivated goal exploration in robots, Robot. Auton. Syst. 61~(1) (2013)
  49--73.
\newblock \href {http://dx.doi.org/10.1016/j.robot.2012.05.008}
  {\path{doi:10.1016/j.robot.2012.05.008}}.

\bibitem{oudeyer2007}
P.-Y. Oudeyer, F.~Kaplan, V.~V. Hafner, Intrinsic motivation systems for
  autonomous mental development, Evolutionary Computation, IEEE Transactions on
  11~(2) (2007) 265--286.

\bibitem{Dillon2013}
M.~R. Dillon, Y.~Huang, E.~S. Spelke,
  \href{http://dx.doi.org/10.1073/pnas.1312640110}{Core foundations of abstract
  geometry.}, Proc Natl Acad Sci U S A 110~(35) (2013) 14191--14195.
\newblock \href {http://dx.doi.org/10.1073/pnas.1312640110}
  {\path{doi:10.1073/pnas.1312640110}}.
\newline\urlprefix\url{http://dx.doi.org/10.1073/pnas.1312640110}

\bibitem{Natale:2007}
L.~Natale, F.~Orabona, G.~Metta, G.~Sandini, Sensorimotor coordination in a
  "baby" robot: learning about objects through grasping, Prog Brain Res 164
  (2007) 403--24.

\bibitem{Fuke:2007}
S.~Fuke, M.~Ogino, M.~Asada, Body image constructed from motor and tactile
  images with visual information., I. J. Humanoid Robotics 4~(2) (2007)
  347--364.

\bibitem{Rolf2014}
M.~Rolf, M.~Asada, Where do goals come from? a generic approach to autonomous
  goal-system development, arXiv preprint arXiv:1410.5557.

\end{thebibliography}

\end{document}